\pdfoutput=1

\documentclass[11pt]{article}
\usepackage[]{ACL2023}

\usepackage{latexsym}

\usepackage{amsmath,amssymb}
\usepackage{txfonts}

\usepackage{caption}

\usepackage[T1]{fontenc}
\usepackage[utf8]{inputenc}
\usepackage{inconsolata}
\usepackage{multirow}
\usepackage{makecell}

\usepackage{graphicx}
\usepackage{pgf}
\usepackage{booktabs}
\usepackage{colortbl}
\usepackage{microtype}
\usepackage{xfp}
\usepackage{twoopt}
\usepackage{xspace}
\usepackage[size=tiny]{todonotes}
\usepackage{subcaption}
\usepackage{tabularx, booktabs}
\usepackage{floatrow}

\usepackage[normalem]{ulem}

\newcommand\pref[2][https://]{%
  \href{#1#2}{\nolinkurl{#2}}%
}

\newcommand\mail[2][mailto:]{%
  \href{#1#2}{\normalcolor\nolinkurl{#2}}%
}

\title{
ChatGPT: 
A Meta-Analysis after 2.5 Months
}

\author{%
Christoph Leiter \and Ran Zhang \and  Yanran Chen \\ {\bf\and Jonas Belouadi \and Daniil Larionov \and Vivian Fresen \and Steffen Eger}\\
\texttt{\{ran.zhang,christoph.leiter,daniil.larionov,jonas.belouadi,steffen.eger\}}\\
\texttt{@uni-bielefeld.de, ychen@techfak.uni-bielefeld.de, V.Fresen@crif.com}
  \AND%
  {\rm Natural Language Learning Group (NLLG)}\\
  Faculty of Technology, Bielefeld University\\
  \pref{nl2g.github.io}
}

\date{}

\makeatletter
\newcommand\scaleinput[1]{%
  \small
  \scalebox{\fpeval{\f@size/9.6}}{\input{#1}} 
}
\newcommand\tightscaleinput[1]{%
  \miniscule%
  \scalebox{\fpeval{\f@size/9.6}}{\input{#1}} 
}

\makeatother

\newcolumntype{s}{>{\hsize=.15\hsize}X}

\begin{document}
\maketitle
\begin{abstract}
ChatGPT, a chatbot developed by OpenAI, has gained widespread popularity and media attention since its release in November 2022. However, little hard evidence is available regarding its perception in various sources. In this paper, we analyze over 300,000 tweets and more than 150 scientific papers to investigate how ChatGPT is perceived and discussed. 
Our findings show that ChatGPT is generally viewed as of high quality, with positive sentiment and emotions of joy dominating in social media. 
Its perception has slightly decreased since its debut, however, with joy decreasing and (negative) surprise on the rise, and it is perceived more negatively in languages other than English. 
In recent scientific papers, ChatGPT is characterized as a great opportunity across various fields including the medical domain, but also as a threat concerning ethics and receives mixed assessments for education. 
 Our comprehensive meta-analysis of ChatGPT's current perception after 2.5 months since its release can contribute to shaping the public debate and informing its future development. We make our data available.\footnote{\url{https://github.com/NL2G/ChatGPTReview}}
\end{abstract}

\section{Introduction}\label{sec:introduction}

\begin{figure*}
    \centering
    \includegraphics[scale=0.5]{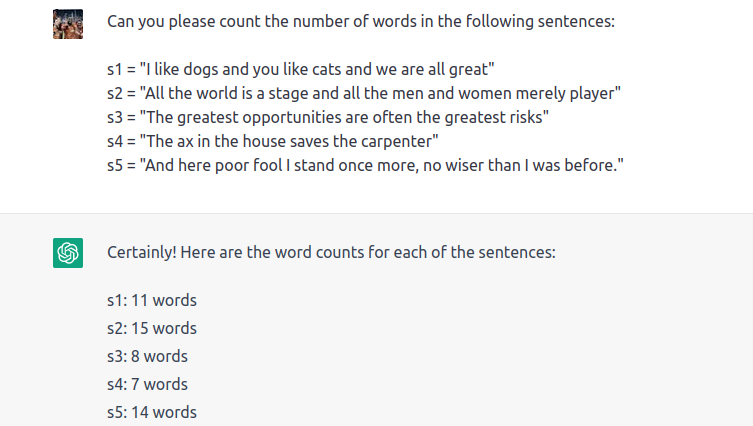}
    \caption{Evidence of counting failures of ChatGPT. The correct answers are 12, 14, 8, 8, 14, of which ChatGPT gets false 3/5 (by an error of one). ChatGPT indicated that it ignored punctuation and quotation marks when counting and also miscounted without any quotation and punctuation marks.}
    \label{figure:counting}
\end{figure*}

ChatGPT\footnote{\url{chat.openai.com/}} --- a chatbot released by OpenAI in November 2022 which can answer questions, write fiction or prose, help debug code, etc.\ --- has seemingly taken the world by storm. Over the course of just a little more than two months, it has attracted more than 100 million subscribers, and 
has been described as the fastest growing web platform ever, leaving behind Instagram, Facebook, Netflix and TikTok\footnote{\url{https://time.com/6253615/chatgpt-fastest-growing/}} \citep{Haque2022ITT}. 
Its qualities have been featured, discussed and praised by popular media,\footnote{   
\url{www.wsj.com/articles/chatgpt-ai-chatbot-app-explained-11675865177}}  
laymen\footnote{\url{www.youtube.com/watch?v=OcXKiTDODFU&t=1151s}} and experts alike. 
On social media, it has (initially) been lauded as ``Artificial General Intelligence'',\footnote{\url{https://twitter.com/MichaelTrazzi/status/1599073962582892546}} 
while more recent assessment hints at limitations and weaknesses e.g.\ regarding its reasoning and mathematical abilities \citep{Borji2023ACA,Frieder2023MathematicalCO} (the authors of this work point out that, as of mid-February 2023, even after 5 updates, ChatGPT can still not accurately count the number of words in a sentence --- see Figure \ref{figure:counting} --- a task primary school children would typically solve with ease.).

However, while there is plenty of anecdotal evidence regarding the perception of ChatGPT, there is little hard evidence via analysis of different sources such as social media and scientific papers published on it. In this paper, we aim to fill this gap. We ask how ChatGPT is viewed from the perspectives of different actors, how its perception has changed over time and which limitations and strengths have been pointed out. We focus specifically on Social Media (Twitter), collecting over 300k tweets, as well as scientific papers from Arxiv and SemanticScholar, analyzing more than 150 papers. 

We find that ChatGPT is overall characterized in different sources as of high quality, with positive sentiment and associated emotions of \emph{joy} dominating. In scientific papers, it is characterized predominantly as a (great) opportunity across various fields, including the medical area and various applications including (scientific) writing as well as for businesses, but also as a threat from an ethical perspective. The assessed impact in the education domain is more mixed, where ChatGPT is viewed both as an opportunity for shifting focus to teaching advanced writing skills \citep{Bishop2023ACW} and for making writing more efficient \citep{Zhai2022ChatGPTUE} but also a threat to academic integrity and fostering dishonesty \citep{Ventayen2023OpenAICG}. 
Its perception has, however, slightly decreased in social media since its debut, with \emph{joy} decreasing and \emph{surprise} on the rise. In addition, in languages other than English, it is perceived with more negative sentiment. 

By providing a comprehensive assessment of its current perception, our paper can contribute to shaping the public debate and informing the future development of ChatGPT.
\section{Analyses}
\subsection{Social Media Analysis}
We aim to acquire 
insights into public opinion and sentiment on ChatGPT and understand public attitudes toward different topics related to ChatGPT. We choose Twitter as our  social media source and collect tweets since the publication date of ChatGPT. The following will introduce the data and the preprocessing steps.  

\paragraph{Dataset} 
We obtain data through the use of a hashtag search tool 
\textit{snscrape},\footnote{\url{https://github.com/JustAnotherArchivist/snscrape}}  
setting our search target as $\#$ChatGPT.  After acquiring the data, we deduplicate all the retweets and remove robots.\footnote{The detection of robots involves the evaluation of two key metrics, the average time between each tweet and the number of total tweets in the examining period. In our analysis, we define the user as a robot account if the average tweet interval between two consecutive tweets is less than 2 hours. We discard 15 such users.} 

\begin{table}
\begin{tabular}{cc}
\toprule
\textbf{Attribute} & \textbf{Detail}\\  \midrule
date range & 2022-11-30 to 2023-02-09\\
number of tweets &  334,808 \\
language counts & 61 \\
English tweets & 228127 \\
number of users &  168,111 \\
\bottomrule
\end{tabular}
\caption{Information of the collected Dataset}
\label{dataset}
\end{table}

Our final dataset contains tweets in the time period from 2022-11-30 18:10:57 to 2023-02-09 17:24:45. The information is summarized in Table \ref{dataset}. We collect over 330k tweets from more than 168k unique user accounts. The average ``age'' over all user accounts is 2,807 
days. On average, each user generates 1.99 tweets over the time period. The dataset contains tweets across 61 languages. Over 68\% of them are in English, other major languages are Japanese (6.4\%), Spanish (5.3\%), French (5.0\%), and German (3.3\%). 
We translate all tweets into English via 
a multi-lingual machine translation model developed by Facebook.\footnote{\url{https://github.com/facebookresearch/fairseq/tree/main/examples/m2m_100}} 

\paragraph{Sentiment Analysis}
We utilize the multi-lingual sentiment classifier from \citet{barbieri-etal-2022-xlm}
to acquire the sentiment label. This XLM-Roberta based language model is trained on 198 million tweets, and finetuned on Twitter sentiment dataset in eight different languages. The model performance on sentiment analysis varies among languages (e.g. the F1-score for Hindi is only 53\%), but the model yields 
acceptable 
results in English with an F1-score of 71\%. 
Thus 
we choose English as our sole input language and collect negative, neutral, and positive sentiments over time (represented as classes 0,1,2, respectively).
\begin{table}
\begin{tabular}{cc}
\toprule
\textbf{Sentiment} & \textbf{Number of tweets}\\  \midrule
Positive & 100,163 \\
Neutral & 174,684 \\
Negative & 59,961 \\
\bottomrule
\end{tabular}
\caption{Sentiment Distribution of all tweets.}
\label{table:sentiment_counts}
\end{table} 
Table \ref{table:sentiment_counts} summarizes the sentiment distribution of all tweets. While the majority of the sentiment is neutral, there is a relatively large proportion of positive sentiment, with
100k 
instances, and a smaller but still notable number of tweets of negative sentiments, with 
60k instances. Table \ref{table:sentiment_sample} provides sample tweets belonging to different sentiment groups.

\begin{table*}
\begin{tabularx}{\textwidth}{Xss}
\toprule
\textbf{Tweet} & \texttt{Sentiment}&\texttt{Topic}\\ 
\midrule
Here we had yet exchanged about the power of open \#KI APIs, now we are immersed in the amazing answers of \#ChatGPT. & 2 & science \& technology\\
I've been playing around with this for a few hours now and I can firmly say that i've never seen anything this developed before. Curious to see where this goes. \#ChatGPT & 2 & diaries \& daily life \\
\midrule
The U.S. company wants to add a filigrane to the texts generated by \#ChatGPT. [url] via @user \#tweetsrevue \#cm \#transfonum & 1 & business \& entrepreneurs \\ 
When you're trying to be productive but the memes keep calling your name.\#TBT \#ChatGPT \#Memes & 1 & diaries \& daily life \\
\midrule
@user I just tested this for myself and it’s TRUE. The platform should be shut down IMMEDIATELY \#chatgpt \#rascist \#woke \#leftwing & 0 & news \& social concern \\ 
I'm starting to think a student used \#ChatGPT for a term paper. If that's the case, the technology isn't ready yet. \#academicchatter & 0  & learning \& educational \\
\bottomrule
\end{tabularx}
\caption{Sample tweets of positive (2), neutral (1), and negative sentiment (0) along with their topic.}
\label{table:sentiment_sample}
\end{table*}

\begin{figure}[htb]
\includegraphics[width=\textwidth]{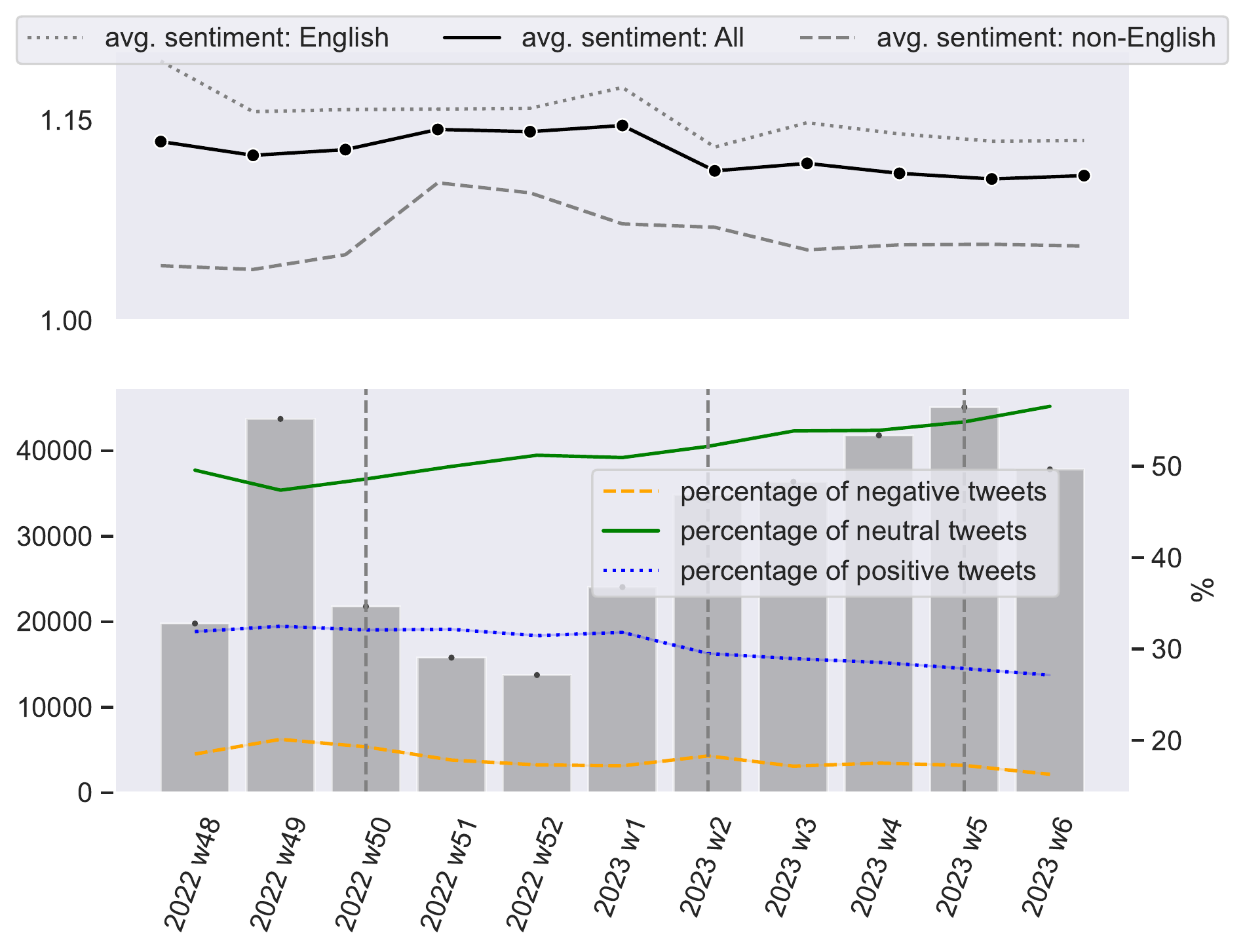}
\caption{Upper: weekly average of  sentiment overall language (solid line), over English tweets (dotted line) and non-English tweets (dashed line). Lower: Tweet counts distribution and sentiment percentage change at weekly level aggregation.}
\label{sentiment_per}
\end{figure}

To examine the sentiment change over time, we plot the weekly average of sentiment and the weekly 
percentage of positive, neutral, and negative tweets in Figure \ref{sentiment_per}. From the upper plot, we observe an overall downward trend of sentiment (black solid line) during the course of ChatGPT's first 2.5 months: 
an initial rise in average sentiment was followed by a decrease from January 2023 onwards. We note, however, that the decline is mild in absolute value: the average sentiment of a tweet decreases from a maximum of about 1.15 to a minimum of 1.10 (which also indicates that the average sentiment of tweets is slightly more positive than neutral). 
We also report the average sentiment of English tweets (dotted line) and non-English tweets(dashed line). Though the absolute difference is small, we can clearly identify the division of sentiment between English and non-English tweets. The difference in sentiment is narrowing over time, but overall tweets in English have a more positive perception of ChatGPT. This suggests that ChatGPT may be better in English, which constituted the majority of its training data; but see also our topic-based analysis below. 

The bar plots in the lower part of the figure represent the count of tweets per week and the line plots show the percentage change of each sentiment class. 
While the percentage of negative tweets is stable over time, the percentage of positive tweets decreases 
and there is a clear increase in tweets with the neutral sentiment. This may indicate that the public view of ChatGPT is becoming more rational after an initial hype of this new ``seemingly omnipotent'' bot. 

During the course of 2.5 months after ChatGPT's debut, OpenAI announced 5 new releases claiming various updates. Our data covers the period of the first three releases on the 15th of December 2022, the 9th of January, and the 3rd of January in 2023. The two latest releases on the 9th of February and the 13th of February are not included in this study.\footnote{\url{https://help.openai.com/en/articles/6825453-chatgpt-release-notes}} The three update time points of ChatGPT are 
depicted as vertical dashed lines in the lower plot of Figure \ref{sentiment_per}.
We can 
observe small short-term increases in sentiment after each new release. 

\paragraph{Sentiment across language and topic} 

\begin{figure}[htb]
\includegraphics[width=\textwidth]{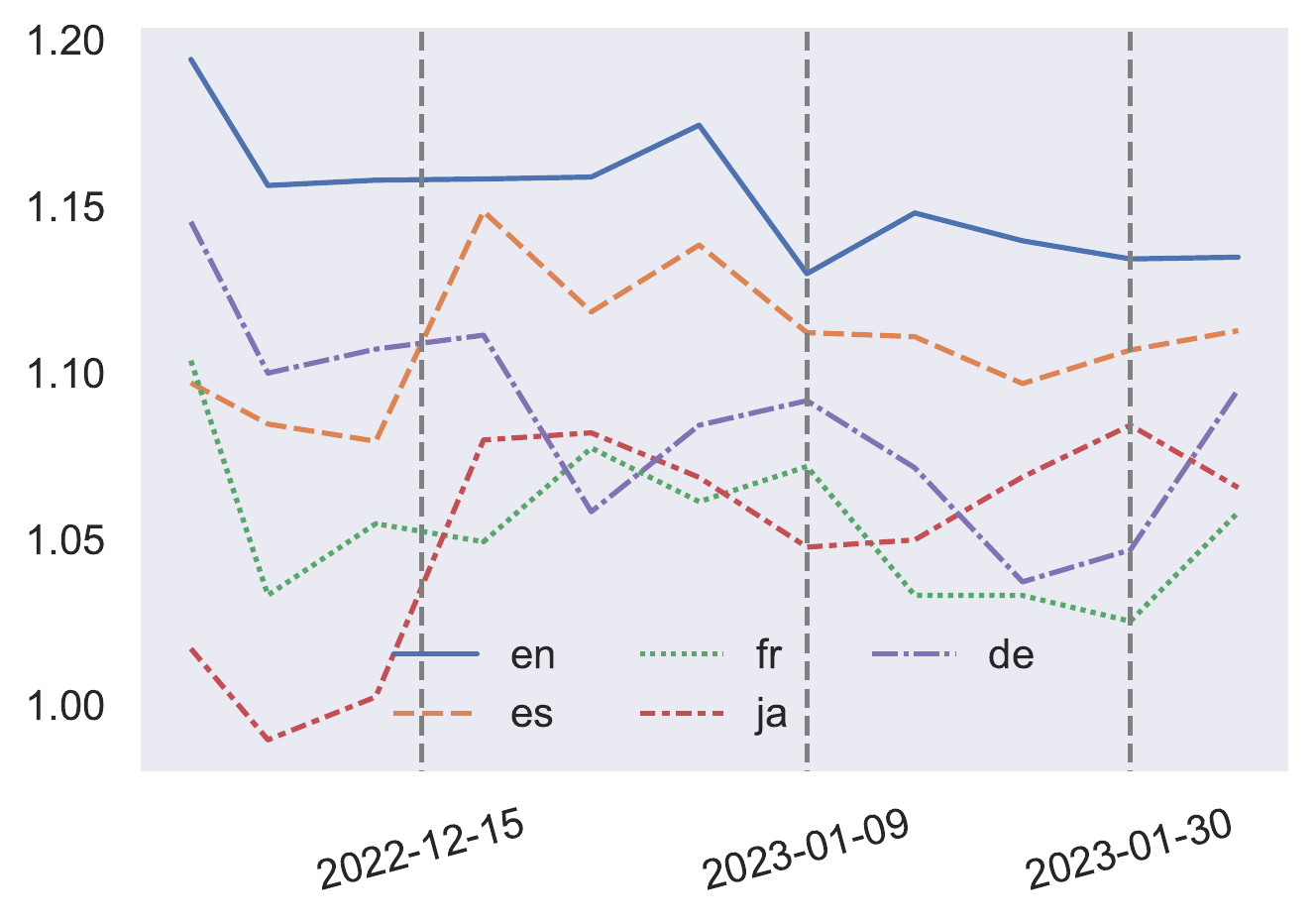}
\caption{Weekly sentiment distribution averaged per language}
\label{sentiment_lang}
\end{figure}
We notice from Figure \ref{sentiment_per} that the sentiments among English and non-English tweets vary. Here we analyze sentiment based on all 5 major languages
in our ChatGPT dataset, namely English (en), Japanese (ja), Spanish (es), French (fr), and German (de). Figure \ref{sentiment_lang} demonstrates the weekly average sentiment of each language over time. As indicated by our previous observation in Figure \ref{sentiment_per}, tweets in English have the most positive view of ChatGPT. 
It is also worth noting that over the time period, the sentiment of English, German, and French tweets are trending downward while Spanish and Japanese tweets 
start from a low point and trend upwards.

\begin{figure}[htb]
\includegraphics[width=\textwidth]{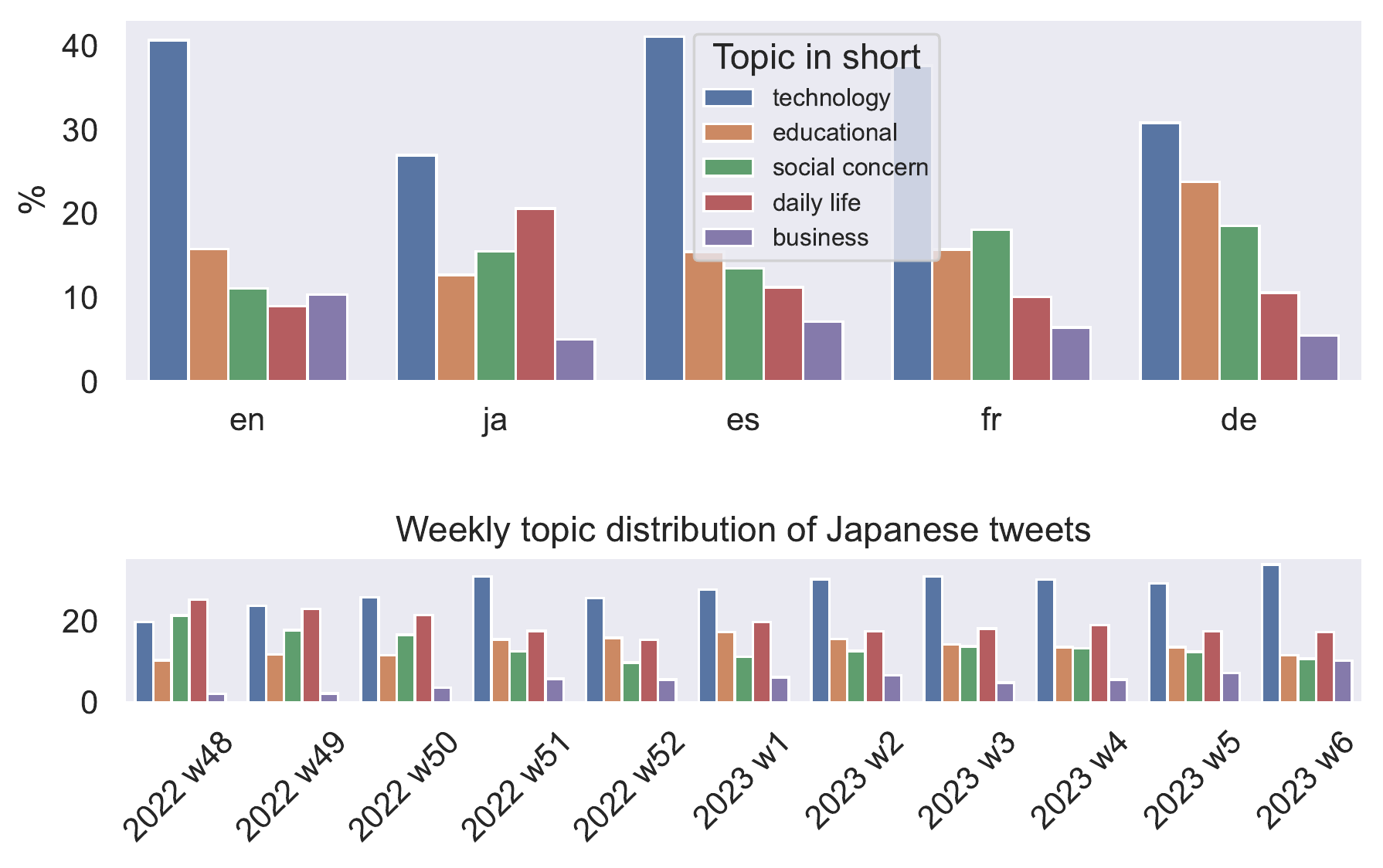}
\caption{Upper: topic  distribution per language. Lower: topic distribution over time for Japanese tweets.}
\label{graph:topic_lang}
\end{figure}

\begin{figure}[htb]
\includegraphics[width=\textwidth]{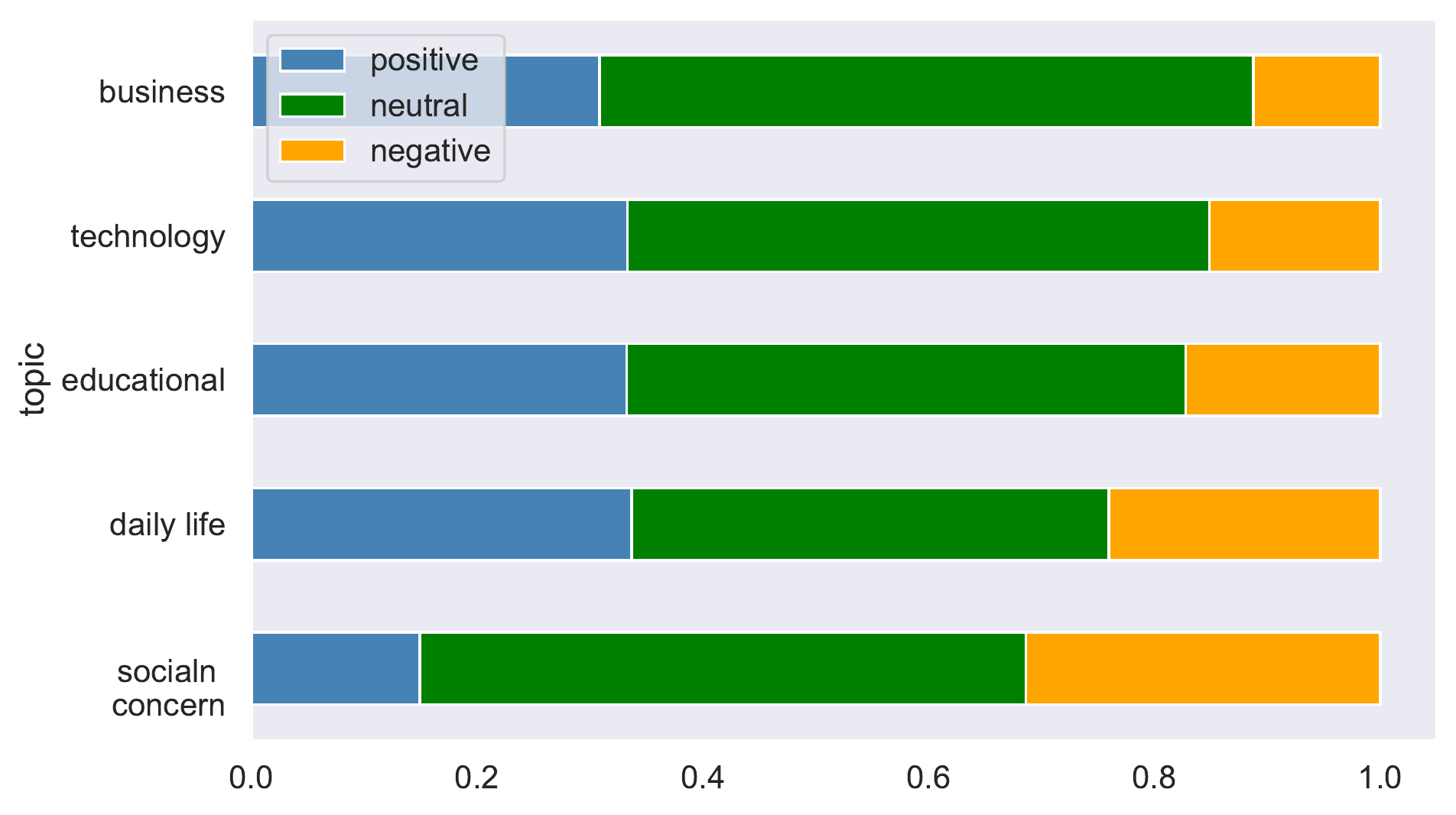}
\caption{Sentiment distribution per topic.}
\label{graph:topic_sent}
\end{figure}

To answer why this is the case, we introduce topic labels into our analysis. 
To do so, 
we utilize the monolingual (English) topic classification model developed by \citet{topic-2022-twitter}. This Roberta-based model is trained on 124 million tweets and finetuned for multi-label topic classification on a corpus of over 11k tweets. The model has 19 classes of topics. 
We only focus on 5 major classes, which cover 86.3\% of tweets in our dataset: 
science \& technology (38.6\%), learning \& educational (15.2\%), news \& social concern (13.0\%), diaries \& daily life (10.2\%), and business \& entrepreneurs (9.3\%). 
The upper plot of Figure \ref{graph:topic_lang} depicts the topic distribution in percentage by different languages. The share of science \& technology topic ranks the highest in all of the 5 languages. However, 
German and French tweets have a relatively higher share of learning \& educational and news \& social concern topics compared to English and Spanish. 
We report the sentiment distribution over different topics in Figure \ref{graph:topic_sent}. From this plot, we notice that the topic business \& entrepreneurs has the lowest proportion of negative tweets while the topic news \& social concern contains the highest proportion of negative tweets. For the other three topics, even though their share of positive tweets are similar, diaries \& daily life topic contains more negative tweets proportionally. 

This observation may 
explain the differences in sentiment distribution among different languages. Compared to other languages, English tweets have the highest proportion of business \& entrepreneurs and science \& technology, both of which contain the lowest share of negative views about ChatGPT. French and German tweets have a similar proportion of  news \& social concern topics, which may result in their slightly less positivity than English tweets, though the three of them have similar overall trends. The case for Japanese and Spanish is unique in terms of the low initial sentiment. The lower plot in Figure \ref{graph:topic_lang}, which shows the topic distribution change over time for Japanese tweets, may explain this phenomenon. We can observe an evident increase in topics concerning business \& entrepreneurs and science \& technology, which contribute more positivity, and a decrease in news \& social concern, which reduces the share of negative tweets. The same explanation may apply to Spanish tweets.  


\paragraph{Aspect of the sentiment}
\par 
To further obtain an understanding of aspects of sentiments in negative and positive tweets, 
we manually annotated and analyzed the sentiment expressed within 40 randomly selected tweets. 
We draw 20 random positive tweets from the period including the last two weeks of 2022 and the first week of 2023, where the general sentiment reaches the peak, and 20 random negative tweets from the second week to the fourth week of 2023, where the general sentiment declines. We are particularly interested in what users find positive/negative about ChatGPT, which in general could relate to many things, e.g., its quality, downtimes, etc. 

\par Based on our analysis of a sample of 20 tweets during the first period, we observed a prevalent positive sentiment towards ChatGPT's ability to generate human-like and concise text. Specifically, 14 out of 20 users reported evident admiration for the model and the text it produced. Users particularly noted the model's capacity to answer complex medical questions, generate rap lyrics and tailor texts to specific contexts. Notably, we also discovered instances where users published tweets that ChatGPT completely generated. 

\par 
As for the randomly selected negative tweets of the second period, 
13 out of the 20 users expressed frustration with the model. These users voiced concerns about potential factual inaccuracies in the generated text and the detectability of the model-generated text. Additionally, a few users expressed ethical concerns, with some expressing worries about biased output or the potential increase in misinformation. Our analysis also revealed that a minority of users expressed concerns over job loss to models like ChatGPT. Overall, these findings suggest that negative sentiment towards ChatGPT was primarily driven by concerns about the model's limitations and its potential impact on society, particularly in generating inaccurate or misleading information.

\par As part of our analysis, we manually evaluated the sentiment categories for the samples analyzed. We found that 25\% (5 out of 20) of the automatically classified sentiment labels were incorrect during the first period. In the second period, we found that 20\% (4 out of 20) of the assigned labels were incorrect. The majority of the misclassified tweets were determined to have a neutral sentiment. Despite these misclassifications, we consider the overall error rate of 22.5\% (9 out of 40) 
acceptable for our use case. 
Especially, errors may cancel out in our aggregated analysis and it is worth pointing out that the main confusions were with the neutral class, not the confusion of negative and positive labels.

\paragraph{Emotion Analysis}

\begin{figure}[!htb]
    \centering
    \includegraphics[width=\textwidth]{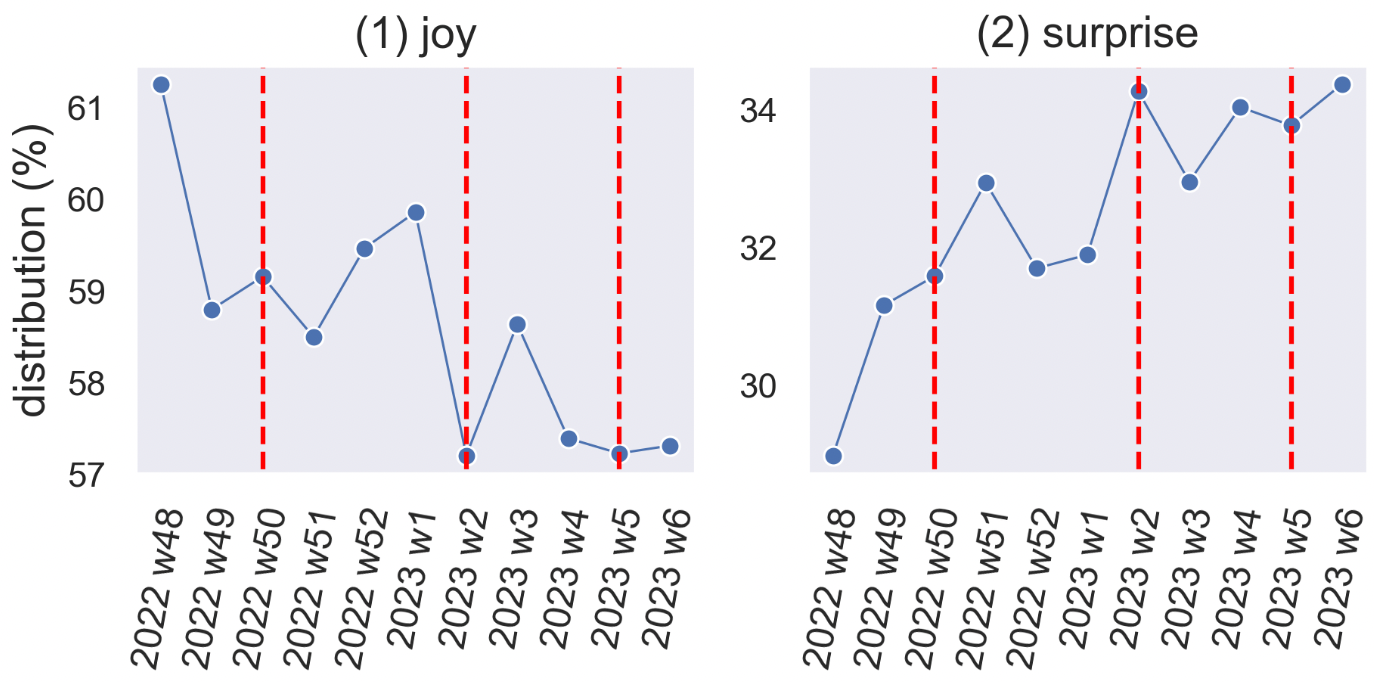}
    \caption{Weekly emotion distribution of (1) \emph{joy} and (2) \emph{surprise} tweets over time. The percentages denote the ratio of \emph{joy}/\emph{surprise} tweets to non-neutral tweets. We mark the update time points with red dashed.}
    \label{fig:emotion}
\end{figure}

In addition to sentiment, we do a more fine-grained analysis based on the emotions of the tweets.
We use the emotion classifier (a BERT base model)
finetuned on the GoEmotions dataset \citep{demszky-etal-2020-goemotions} that contains texts from Reddit and their emotion labels based on Ekman's taxonomy \citep{ekman1992there}
to categorize the translated English tweets 
into 7 dimensions:
\emph{joy, surprise, anger, sadness, fear, disgust} and \emph{neutral}.\footnote{
We use it to predict a single label for each tweet, despite that it is a multilabel classifier (\url{https://huggingface.co/monologg/bert-base-cased-goemotions-ekman}).}
Among all 334,808 tweets, the great majority 
are labeled as \emph{neutral} ($\sim$70\%), followed by the ones classified as \emph{joy} (17.6\%) and \emph{surprise} (9.8\%); the tweets classified as the remaining 4 emotions compose only 2.7\% of the whole dataset.

We demonstrate the weekly changes in the emotion distribution of 
\emph{joy} and \emph{surprise} tweets
in Figure \ref{fig:emotion}. 
Here we only show the percentage distribution denoting the ratio of the tweets classified as a specific emotion to all tweets with emotions (i.e., the tweets which are not labeled as \emph{neutral}). 
We observe that 
the percentage of \emph{joy} tweets generally decreases after the release, though it rises to some degree after each update, indicating that the users have less fun with ChatGPT over time.
On the other hand, the percentage of 
\emph{surprise} tweets is overall in an uptrend with slight declines between the update time points. 

To gain more insights, 
we manually analyze five randomly selected tweets  per emotion category for the release and each of the three update dates.\footnote{We draw the random sample from the tweets posted two days after each date considering the difference in time zones.} 
Here, we focus on the \emph{joy} and \emph{surprise} tweets, as they dominate in the tweets with emotions; additionally, we also include an analysis of \emph{fear} tweets, because of the observed peak in their distribution trend at the first two update time points, which we believe could 
provide more insight into the users' concerns across different updates. We collect a total of 60 tweets for manual analysis (5 tweets $\times$ 4 dates $\times$ 3 emotions); we show one sample for each emotion in Table \ref{table:emotion_sample}.

\begin{table*}[!htb]
\begin{tabularx}{\textwidth}{Xss}
\toprule
\textbf{Tweet} & \textbf{Emotion} \\
\midrule
\#ChatGPT is so excellent, so fun and I touch it every day, but I’m looking for a way to use it every day. & \emph{joy}\\ \midrule
Wow \! just wow, Just asked \#ChatGPT to write a vision statement for \#precisiononcology [emoji] \! \#AI is [emoji] @user @user @user @user [url] & \emph{surprise} 
\\ \midrule
ChatGPT is taking over the internet, and I am afraid, the world for good! \#ChatGPT & \emph{fear}\\ \bottomrule
\end{tabularx}
\caption{Sample automatically labeled \emph{joy}, \emph{surprise} and \emph{fear} tweets. We mask the user and url information in the table.}
\label{table:emotion_sample}
 \end{table*}
\paragraph{joy}
Our own annotation suggests that 2 out of 20 tweets were misclassified to this category. Among the 18 tweets correctly classified, 
12 tweets directly expressed admiration or reported positive interactions with ChatGPT such as successfully performing generation tasks or acquiring answers, 1 tweet conveyed a positive outlook of AI in NFT (Non-Fungible Token) and game production, and 5 tweets expressed joy which is, however, not (directly) related to ChatGPT. Interestingly, even though 3 
tweets did not pertain directly to ChatGPT, they expressed delight in a talk, post, or interview about ChatGPT, and all of them were posted after the second or third updates.

\paragraph{fear}
1 out of 20 tweets was found to be misclassified, and 1 tweet expressed fear but was unrelated to ChatGPT. Among the remaining 18 tweets, 9 expressed 
scariness 
of ChatGPT because of its strong capability, 
1 user argued that google should be scared of ChatGPT, and the rest 8 tweets reported various concerns including providing wrong/malicious information, job loss and the unethical use of ChatGPT. It is noteworthy that 7 out of the 8 tweets demonstrating concerns were published after the second or third updates. 

\paragraph{surprise}
Among the sampled 
tweets, 
we found that more misclassified tweets may exist compared to the other two categories; the model tends to classify the sentences with question marks as surprise. It is also 
a challenging task for humans to identify ``surprise'' in a short sentence, as this emotion may involve different cognitive and perceptual processes. 
Moreover, ``surprise'' 
could have both negative and positive connotation. 
Hence, we do a four-way manual sentiment annotation for the \emph{surprise}
tweets: positive, negative, mixed and unrelated.
12 out of 20 tweets were found to be correctly classified as surprise, among which
6 tweets conveyed positive surprise due to ChatGPT's impressive performance, 2 tweets expressed negative surprise in terms of providing inaccurate information and prejudice 
against AI, and 4 tweets expressed positive surprise about ChatGPT but with negative concerns regarding unethical uses. 
We further consider all tweets expressing surprise before and after the 2nd update. Before the 2nd update, there were 1.13 times more positive sentiment surprise tweets than negative ones (4942 vs.\ 4372); after the second update, the ratio is roughly equal (5065 vs.\ 5093). 

The decrease in \emph{joy} tweets and the increase in negative \emph{surprise} tweets over time --- even though on relatively small levels --- indicates a more nuanced and rational assessment of ChatGPT over time, similar to the overall decline of positive sentiment over time found in our initial sentiment analysis. We still notice that, apart from \emph{neutral}, \emph{joy} is the most frequently expressed emotion for tweets relating to \#ChatGPT in our sample.

\begin{figure*}
  \begin{subfigure}{0.49\textwidth}
  \centering
    \includegraphics[width=\textwidth]{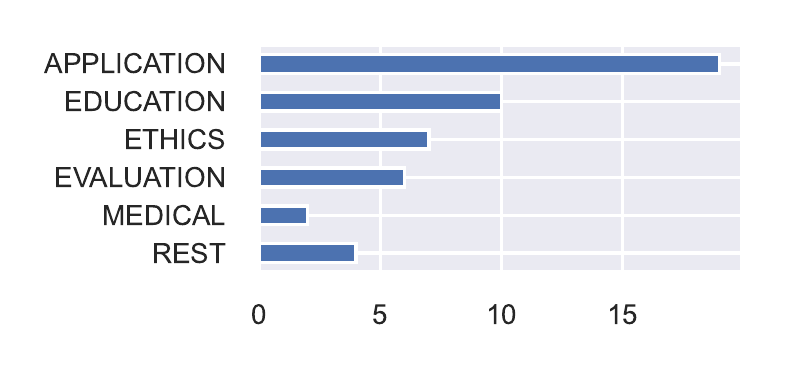}
\caption{Arxiv}
\label{topic_arxiv}
  \end{subfigure}
  \begin{subfigure}{0.49\textwidth}
    \centering
    \includegraphics[width=\textwidth]{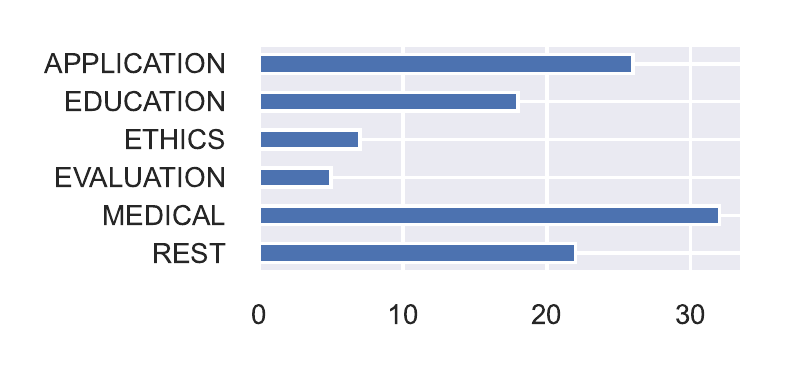}
\caption{SemanticScholar}
    \label{topic_scholar}
  \end{subfigure}
  \caption{Topics of papers found on Arxiv and SemanticScholar that include ChatGPT in their title or abstract. \textit{Ethics} also comprises papers addressing AI regulations and cyber scurity. \textit{Evaluation} denotes papers that evaluate ChatGPT with respect to biases or more than a single domain.}
  \label{topic_overview}
\end{figure*}

\begin{figure*}
  \begin{subfigure}{0.49\textwidth}
  \centering
    \includegraphics[width=\textwidth]{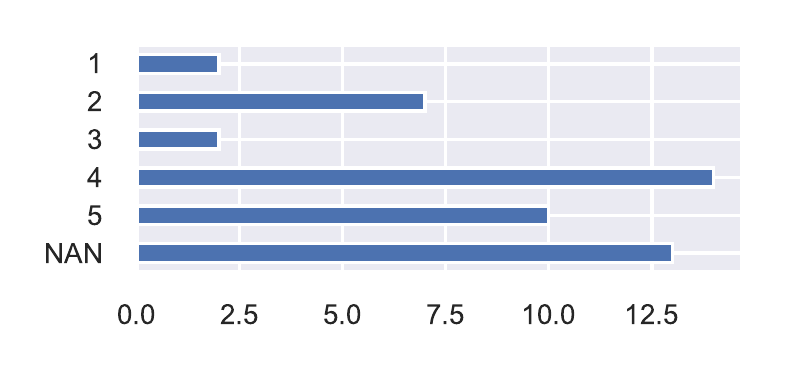}
\caption{Arxiv}
\label{performance_arxiv}
  \end{subfigure}
  \begin{subfigure}{0.49\textwidth}
    \centering
    \includegraphics[width=\textwidth]{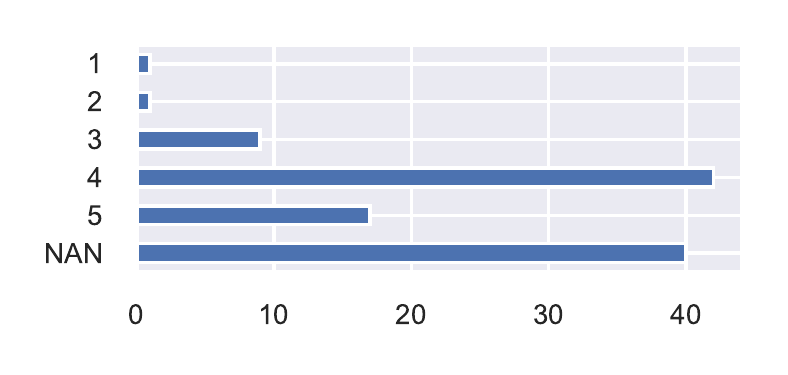}
\caption{SemanticScholar}
\label{performance_scholar}
  \end{subfigure}
  \caption{Performance quality in papers found on Arxiv and SemanticScholar that include ChatGPT in their title or abstract. On a scale of 1 (bad performance) to 5 (good performance), this indicates how the performance of ChatGPT is described in the papers' titles/abstracts. \emph{NAN} indicates that no performance sentiment is given.}
  \label{performance_overview}
\end{figure*}

\begin{figure*}
  \begin{subfigure}{0.49\textwidth}
  \centering
    \includegraphics[width=\textwidth]{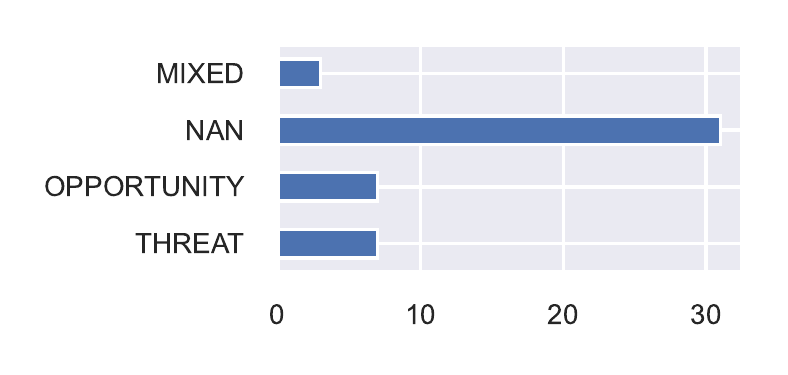}
\caption{Arxiv}
\label{social_arxiv}
  \end{subfigure}
  \begin{subfigure}{0.49\textwidth}
    \centering
    \includegraphics[width=\textwidth]{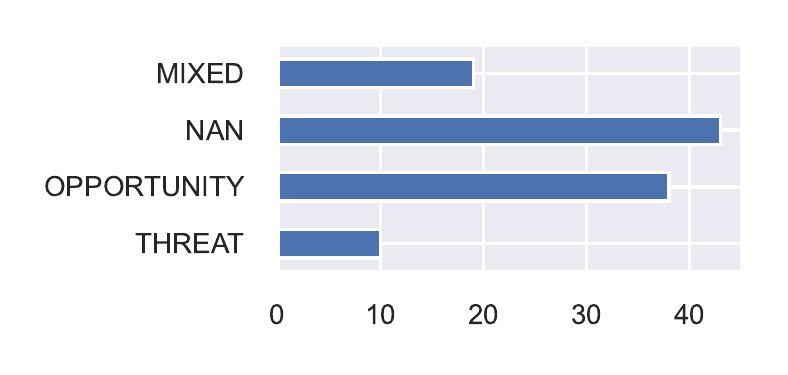}
\caption{SemanticScholar}
\label{social_scholar}
  \end{subfigure}
  \caption{Social impact in papers found on Arxiv and SemanticScholar that include ChatGPT in their title or abstract. The labels indicate (based on abstract and title) which effect the authors believe ChatGPT will have on the social good. \textit{NAN} indicates that no social sentiment is given.}
  \label{social_overview}
\end{figure*}

\begin{figure}[htb]
\includegraphics[width=\textwidth]{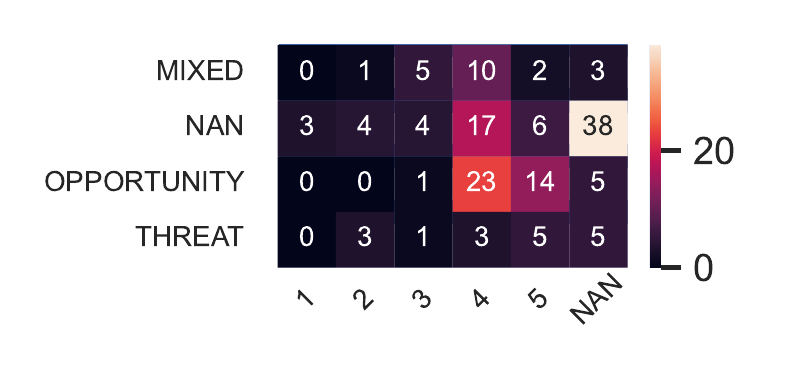}
\caption{Heatmap of performance quality and social impact of papers from Arxiv and SemanticScholar. On a scale of 1 (bad performance) to 5 (good performance), the performance quality indicates how the performance of ChatGPT is described in the papers' titles/abstracts. NAN indicates that no performance quality is given. The social impact indicates (based on abstract and title), which effect the authors belief ChatGPT will have on the social good. \textit{NAN} indicates that no social impact is given.}
\label{performance_social}
\end{figure}

\begin{figure}[htb]
\includegraphics[width=\textwidth]{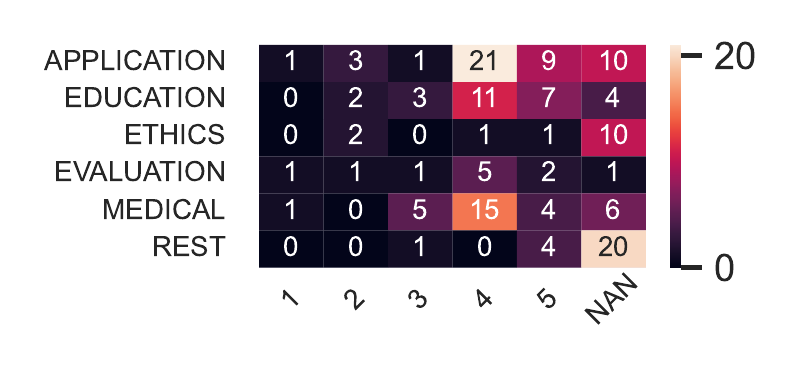}
\caption{Heatmap of performance quality and topic of papers from Arxiv and non-Arxiv papers retrieved from SemanticScholar. On a scale of 1 (bad performance) to 5 (good performance), the performance quality indicates how the performance of ChatGPT is described in the papers' titles/abstracts. NAN indicates that no performance quality is given. \textit{Ethics} also comprises papers addressing AI regulations and cyber scurity. \textit{Evaluation} denotes papers that evaluate ChatGPT with respect to 
multiple aspects. 
}
\label{performance_topic}
\end{figure}

\begin{figure}[htb]
\includegraphics[width=\textwidth]{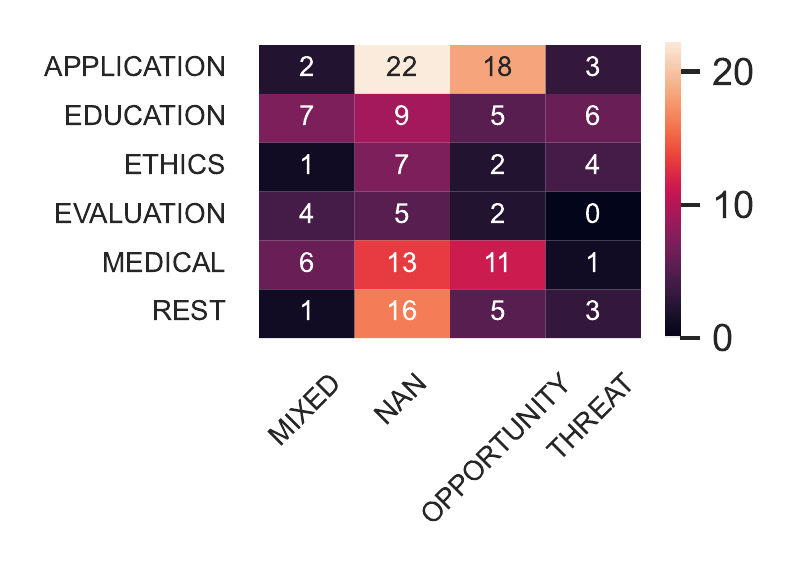}
\caption{Heatmap of social impact and topic of papers from Arxiv and non-Arxiv papers retrieved from SemanticScholar. The social impact indicates (based on abstract and title), which effect the authors belief ChatGPT will have on the social good. \textit{NAN} indicates that no social impact is described. \textit{Ethics} also comprises papers addressing AI regulations and cyber scurity. \textit{Evaluation} denotes papers that evaluate ChatGPT with respect to biases or more than a single domain.}
\label{social_topic}
\end{figure}

\begin{figure}[htb]
\includegraphics[width=\textwidth]{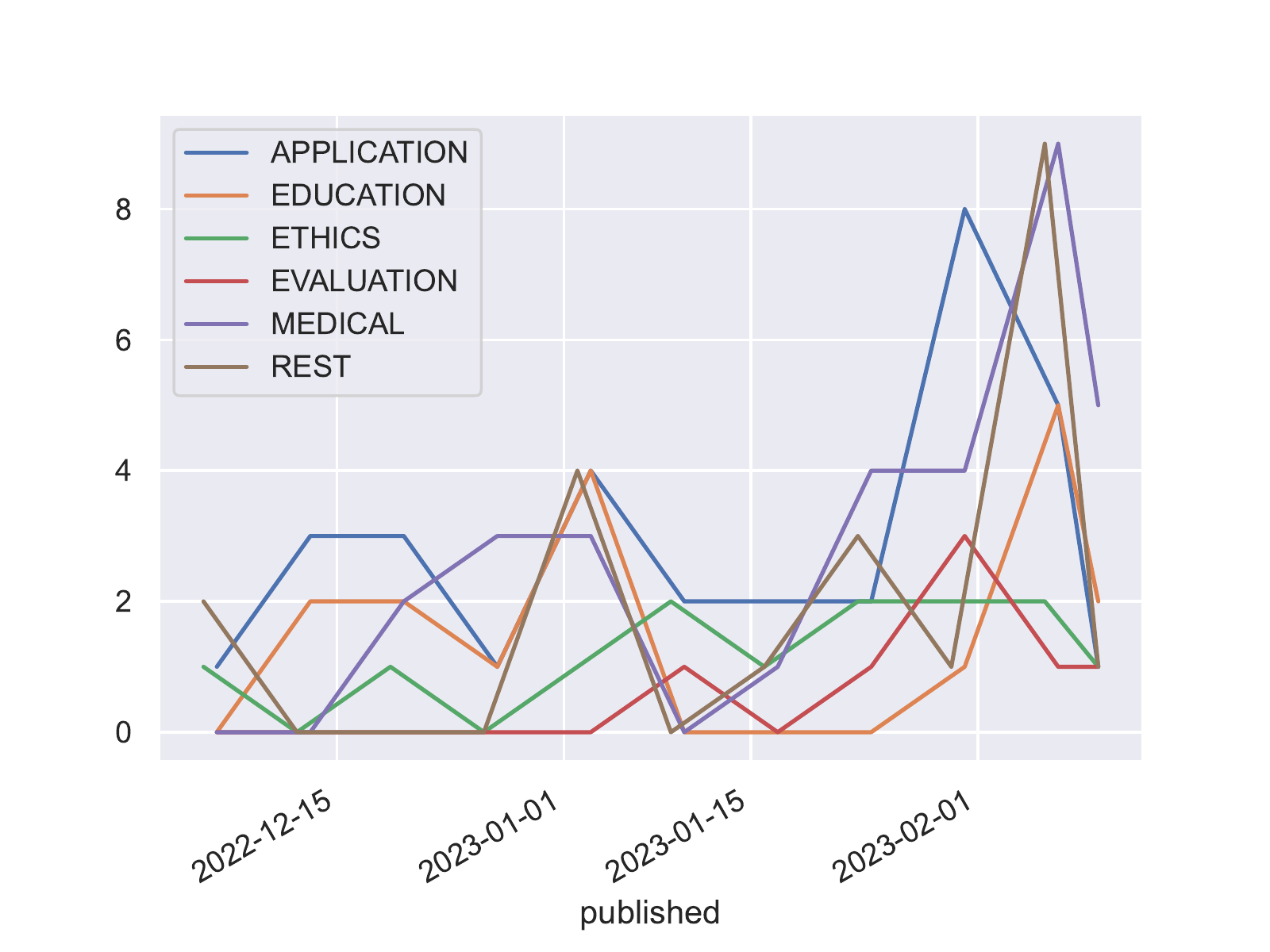}
\caption{Paper topics per 7 days. The downtrend in the last days is an artefact of the time aggregation.}
\label{time_topic}
\end{figure}

\subsection{Arxiv \& SemanticScholar}

\begin{table*}
{\small
\begin{tabularx}{\textwidth}{Xsss}
\toprule
\textbf{Abstract} & \texttt{Topic} & \texttt{Quality} & \texttt{Impact}\\
\midrule
This study evaluated the ability of ChatGPT, a recently developed artificial
intelligence (AI) agent, 
to perform high-level cognitive tasks and produce text
that is indistinguishable from human-generated text. This capacity raises
concerns about the potential use of \uline{ChatGPT as a tool for academic misconduct
in online exams}. \textbf{The study found that ChatGPT is capable of exhibiting critical
thinking skills and generating highly realistic text with minimal input, making
it a potential \uline{threat} to the integrity of online exams, particularly in
tertiary education settings where such exams are becoming more prevalent.}
Returning to invigilated and oral exams could form part of the solution, while
using advanced proctoring techniques and AI-text output detectors may be
effective in addressing this issue, they are not likely to be foolproof
solutions. Further research is needed to fully understand the implications of
large language models like ChatGPT and to devise strategies for combating the
risk of cheating using these tools. It is crucial for educators and
institutions to be aware of the possibility of ChatGPT being used for cheating
and to investigate measures to address it in order to maintain the fairness and
validity of online exams for all students. \citep{Susnjak2022ChatGPTTE} 
& Education & 5 & Threat\\
\midrule
This report provides a preliminary evaluation of ChatGPT for machine
translation, including translation prompt, multilingual translation, and
translation robustness. We adopt the prompts advised by ChatGPT to trigger its
translation ability and find that the candidate prompts generally work well and
show minor performance differences. By evaluating on a number of benchmark test
sets, \textbf{we find that ChatGPT performs competitively with commercial translation
products (e.g., Google Translate) on high-resource European languages but lags
behind significantly on low-resource or distant languages. For distant
languages, we explore an interesting strategy named {pivot~prompting}
that asks ChatGPT to translate the source sentence into a high-resource pivot
language before into the target language, which improves the translation
performance significantly. As for the translation robustness, ChatGPT does not
perform as well as the commercial systems on biomedical abstracts or Reddit
comments but is potentially a good translator for spoken language.} \citep{Jiao2023IsCA}
& Application & 3 & NAN\\ \midrule
Of particular interest to educators, \textbf{an exploration of what new language-generation software} \textbf{does (and does not) do well}. Argues that the new language-generation models make instruction in writing mechanics irrelevant, and that \uline{educators should shift to teaching only the more advanced writing skills that reflect and advance critical thinking}. The difference between mechanical and advanced writing is illustrated through a ``Socratic Dialogue'' with ChatGPT. Appropriate for classroom discussion at High School, College, Professional, and PhD levels. \citep{Bishop2023ACW} & Education & 4 & Opportunity \\ \midrule
We investigate the mathematical capabilities of ChatGPT by testing it on
publicly available datasets, as well as hand-crafted ones, and measuring its
performance against other models trained on a mathematical corpus, such as
Minerva. We also test whether ChatGPT can be a useful assistant to professional
mathematicians by emulating various use cases that come up in the daily
professional activities of mathematicians (question answering, theorem
searching). In contrast to formal mathematics, where large databases of formal
proofs are available (e.g., the Lean Mathematical Library), current datasets of
natural-language mathematics, used to benchmark language models, only cover
elementary mathematics. We address this issue by introducing a new dataset:
GHOSTS. It is the first natural-language dataset made and curated by working
researchers in mathematics that (1) aims to cover graduate-level mathematics
and (2) provides a holistic overview of the mathematical capabilities of
language models. We benchmark ChatGPT on GHOSTS and evaluate performance
against fine-grained criteria. We make this new dataset publicly available to
assist a community-driven comparison of ChatGPT with (future) large language
models in terms of advanced mathematical comprehension. We conclude that
contrary to many positive reports in the media (a potential case of selection
bias), \textbf{ChatGPT's mathematical abilities are significantly below those of an
average mathematics graduate student}. Our results show that ChatGPT often
understands the question \textbf{but fails to provide correct solutions. Hence, if your
goal is to use it to pass a university exam, you would be better off copying
from your average peer}! \citep{Frieder2023MathematicalCO}
& Evaluation & 1 & NAN 
\\
\bottomrule
\end{tabularx}
}
\caption{Sample annotated abstracts from Arxiv (top 2) and SemanticScholar (bottom) along with annotation dimensions topic, quality (attributed to ChatGPT) and impact on society.}
\label{table:arxiv}
\end{table*}

Given the limited time frame of ChatGPT's availability, a substantial portion of potentially relevant papers on it are not yet available in officially published form. 
Thus, we focus our analysis on two sources of information: (1) preprints from Arxiv, which may or may not have already been published; and (2) non-Arxiv papers identified through SemanticScholar. The Arxiv preprints primarily comprise computer science and similar ``hard science'' disciplines. 
Arxiv papers may 
represent the cutting edge of research in these fields \citep{Eger2018PredictingRT}. On the other hand, non-Arxiv SemanticScholar papers encompass a broad range of academic disciplines, including the humanities and social sciences. 
We do not automatically classify papers but resort to manual annotation, which is feasible given that there are only $\sim$150 papers in our dataset, see Table \ref{table:elementary_stats}. 

\begin{table}
\begin{tabular}{cc}
\toprule
\textbf{Source} & \textbf{Number of instances}\\  \midrule
Arxiv & 48 \\
SemanticScholar & 104 \\
\bottomrule
\end{tabular}
\caption{Number and source of scientific papers examined. Here, SemanticScholar comprises only non-Arxiv papers.}
\label{table:elementary_stats}
\end{table}

\paragraph{Annotation scheme}
We classify papers along three dimensions: 
\begin{itemize}
\item[(i)] their \textbf{\texttt{quality}} assessment of ChatGPT. The range is 1-5, where 1 indicates very low and 5 very high assessment, 3 is neutral. NAN indicates that the paper does not discuss the quality of ChatGPT.
\item[(ii)] their \textbf{\texttt{topic}}. After checking the papers and some discussion, we decided on six different topics. These are \emph{Ethics} (which includes biases, fairness, security, etc.), \emph{Education}, \emph{Evaluation} (which includes reasoning, arithmetic, logic problems, etc.\ on which ChatGPT is evaluated), \emph{Medical}, \emph{Application} (which includes writing assistance or using ChatGPT in downstream tasks such as argument mining, coding, etc.) and \emph{Rest}. We note that a given paper could typically be classified into multiple classes, but we are interested in the dominant class. 
\item[(iii)] their \textbf{\texttt{impact}} on society. We distinguish \emph{Opportunity}, \emph{Threat}, \emph{Mixed} (when a paper highlights both risks and opportunities) or \emph{NAN} (when the paper does not discuss this aspect). 
\end{itemize}

Example annotations are shown in Table \ref{table:arxiv}.

\paragraph{Annotation outcomes} Four co-authors of this paper (three male PhD students and one male faculty member) initially annotated 10 papers on all three dimensions independently without guidelines. Agreements were low across all dimensions. After a discussion of disagreements, we devised guidelines for subsequent annotation of 10 further papers. This included (among others) 
to only look at paper abstracts for classification, as the annotation process would otherwise be too time-consuming, and which labels to prioritize in ambiguous cases. Abstracts are a good compromise because abstracts are (highly condensed) summaries of scientific papers, containing their main message. This time, agreements were high: the kappa agreement is 0.63 on average across all pairs of annotators for \texttt{topic}, 0.70 for \texttt{impact} and 0.80 Spearman for \texttt{quality}, averaged across annotators. In total we annotated 48 papers from Arxiv and 104 additional papers from SemanticScholar.

\paragraph{Analysis}
Figure \ref{topic_overview} shows the topic distributions for the Arxiv papers and the papers from SemanticScholar. The main topics we classified for the Arxiv papers are \textit{Education} and the \textit{Application} in various use cases. Only few papers were classified as \textit{Medical}. 
Conversely, SemanticScholar papers are most frequently classified as 
\textit{Medical} and \textit{Rest}. 
This indicates that \textit{Medical} is of great concern in more applied scientific fields not covered by Arxiv papers.   
Further, Figure \ref{performance_overview} shows the distributions of 
quality 
labels we annotated. The labels 4 and 5 have high numbers of occurrences, i.e., many papers report a strong performance of ChatGPT. 
Figures \ref{social_arxiv} and \ref{social_scholar} show the distributions for our annotations of the social impact. 
If a social impact sentiment is provided, 
ChatGPT is most frequently described as an opportunity. For Arxiv, the number of papers which see ChatGPT as an opportunity is the same number as papers that see it as a threat. 

In the second part of the analysis we consider the annotations from Arxiv and SemanticScholar together. Figure \ref{performance_social} displays the intersection of \textit{performance quality} and \textit{social impact}. It shows that authors who report a high performance quality for ChatGPT ($4$/$5$) in most cases also believe that it will have a positive social impact. Also, there is a high number of papers which reported no performance quality or social impact (NAN). Papers that report a low performance quality ($1$/$2$) either state no social impact or perceive it as \textit{Mixed} or a \textit{Threat}, but not as \textit{Opportunity}. Figure \ref{performance_topic} shows the intersection between \textit{performance quality} and \textit{topic}. For every topic, the majority of papers describe a high performance quality of ChatGPT. Also, most papers that report low quality are found for \textit{Application} and \textit{Education}. Lastly, Figure \ref{social_topic} presents the intersection of \textit{topic} and \textit{social impact}. Here, papers in the categories \textit{Application}, \textit{Medical} and \textit{Rest} mostly describe ChatGPT as an opportunity for society. For \textit{Education}, the number of papers that see ChatGPT as a threat is almost equal to the number of those that view it as opportunity. For \textit{Evaluation}, a comparably high number of abstracts articulate mixed sentiments towards the social impact. Finally, in the \textit{Ethics} category, ChatGPT is mostly seen as a threat. 

We also consider the development of each annotated category over time, using all considered papers from Arxiv and those of SemanticScholar that have an attached publication date. Overall, the amount of papers that is published every week is increasing, highlighting the current importance of the topic. Compared to the Twitter data, the sample size of papers is small, hence, other trends are difficult to reliably describe. In Figure \ref{time_topic}, we show that the topics \textit{Evaluation} and \textit{Ethics} have not been considered as a main topic in most early papers of December 2022. Further, the amount of papers in \textit{Medical} and \textit{Rest} increases especially since the beginning of February, showing the newly gained, widespread recognition of ChatGPT in many areas outside the NLP community. 

To conclude, the analysis of papers exemplifies the explosive attention ChatGPT is getting. They mostly see ChatGPT as an opportunity for society and praise its performance. Threats perceived in \textit{Education} and \textit{Ethics} could for example be linked to concerns about plagiarism \cite[e.g.]{https://doi.org/10.48550/arxiv.2212.11661}.

\begin{table*}
\begin{tabular}{cc}
\toprule
\textbf{Category} & \textbf{Labels}\\  \midrule
\texttt{Topic} & Ethics, Education, Evaluation, Medical, Application, Rest\\
\texttt{Quality} & 0 ($=$ NAN), 1,2,3,4,5 \\
\texttt{Impact} & Threat, Opportunity, Mixed, NAN \\
\bottomrule
\end{tabular}
\caption{Annotation categories for Arxiv and SemanticScholar papers along with their labels. Quality refers to the assessment of papers with regard to ChatGPT; 5 indicates that the papers attest it very high quality and 1 indicates very low quality. Impact refers to whether papers describe ChatGPT as a Threat or Opportunity (for society).}
\label{table:annotation_labels}
\end{table*}



\subsection{Other sources}\label{sec:analysis}
Up to now, 
we analyzed the public opinion of the ChatGPT model by
analyzing arXiv/SemanticScholar papers and Twitter for sentiment.

However, it is important to note that there are other resources that can
provide valuable insights into the model. One such resource is GitHub repositories,
which contain a wealth of information about the ChatGPT model. This includes
third-party libraries that can be used to programmatically leverage or even
enhance the functionality of the
model,\footnote{%
\url{https://github.com/stars/acheong08/lists/awesome-chatgpt}\\
\url{https://github.com/saharmor/awesome-chatgpt}\\
\url{https://github.com/humanloop/awesome-chatgpt}
} as well as lists of prompts that can be used to test its
abilities.\footnote{%
\url{https://github.com/f/awesome-chatgpt-prompts}\\
\url{https://chatgpt.getlaunchlist.com}\\
\url{https://promptbase.com/marketplace}
}

Other valuable resource are blog posts and the discussion of failure cases,\footnote{%
\url{https://github.com/giuven95/chatgpt-failures}\\
\url{https://docs.google.com/spreadsheets/d/1kDSERnROv5FgHbVN8z_bXH9gak2IXRtoqz0nwhrviCw}}
 which can help us
understand the limitations of the model and how they can be addressed. These
resources provide important feedback to the developers and can inform future
development efforts, ensuring that the ChatGPT model continues to evolve and
improve.

We constructed a small dataset (50 entries) of such online resources and enlisted two coworkers to annotate their sentiment. Our analysis of these resources (see Table~\ref{tab:web-sentiment}) reveals that shared prompts lists, as well as other GitHub repositories exhibit  overwhelmingly positive sentiment, while blog posts display a mix of positive, neutral, and negative sentiments in nearly equal proportions.
\begin{table*}[]
    \centering
    \begin{tabular}{lrrr}
    \toprule
    \textbf{Type} & \textbf{Positive} & \textbf{Neutral} & \textbf{Negative}\\
    \midrule
    Prompt Sharing Sites   & 100\% &  0\% &   0\%\\
    GitHub Repositories    &  92\% &  8\% &   0\%\\
    Blog Posts             &  22\% & 44\% &  33\%\\
    Lists of Failure Cases &   0\% &  0\% & 100\%\\
    \bottomrule
    \end{tabular}
    \caption{Observed sentiment found in various independent resources around the web.}
    \label{tab:web-sentiment}
\end{table*}
We further observed that lists of failure cases showed the poorest overall sentiment, a finding
which intuitively makes sense.
Failure cases often involve math problems,\footnote{\url{https://twitter.com/GaryMarcus/status/1610793320279863297}} a domain where ChatGPT frequently provides confidently incorrect answers. However, our findings were not entirely consistent, as some positive blog posts suggest that ChatGPT performs well in symbolic execution of code,\footnote{\url{https://www.engraved.blog/building-a-virtual-machine-inside}} indicating that the issue may lie in prompt tuning rather than ChatGPT's general capabilities, i.e., ChatpGPT can handle math problems better when they are formulated as programs, not prose.
Neutral\footnote{\url{https://thezvi.substack.com/p/jailbreaking-the-chatgpt-on-release}} and negative\footnote{\url{https://davidgolumbia.medium.com/chatgpt-should-not-exist-aab0867abace}} blog posts tend to focus less on the quality of ChatGPT's outputs and more on concerns related to OpenAI's restrictions or potential negative social impacts.
\section{Related work}

The two most closely related works are \citet{Haque2022ITT,Borji2023ACA}. \citet{Haque2022ITT} study twitter reception of ChatGPT after about 2 weeks, finding that the majority of tweets have been overwhelmingly positive in this early period. 
They have much smaller samples which they manually annotate and use unsupervised topic modeling to determine topics. They also do not look at scientific papers, but only at social media posts. 
\citet{Borji2023ACA} presents a catalogue of failure cases of ChatGPT relating to reasoning, logic, arithmetic, factuality, bias and discrimination, etc. The failure cases are based on selected examples mostly from social media.  
\citet{bowman2022dangers,beese2022detecting} discuss the increase of negative papers over time using NLP tools, which is also related to our study. In contrast to their work, we only discuss very recent trends over the last months; our methodological setup is also very different. 
\section{Conclusion}\label{sec:conclusion}

In this paper, we conducted a comprehensive analysis of the perception of ChatGPT, a chatbot released by OpenAI in November 2022 that has attracted over 100 million subscribers in only two months. We analyzed over 300k tweets and more than 150 scientific papers to understand how ChatGPT is viewed from different perspectives, how its perception has changed over time, and what its strengths and limitations are. We found that ChatGPT is generally perceived positively, with high quality, and associated emotions of joy dominating. However, its perception has slightly decreased since its debut, and in languages other than English, it is perceived with more negative sentiment. Moreover, while ChatGPT is viewed as a great opportunity across various scientific fields, including the medical domain, it is also seen as a threat from an ethical perspective and in the education domain. Our findings contribute to shaping the public debate and informing the future development of ChatGPT.

Future work should 
investigate developments over longer stretches of time, consider popularity of tweets and papers (via likes and citations), investigate more dimensions besides sentiment and emotion and look at the expertise of social media actors and their geographic and demographic distribution. 
Finally, as language models like ChatGPT continue to evolve and gain more capabilities, future research can assess their real (rather than anticipated) impact on society, including their potential to exacerbate and mitigate existing inequalities and biases.

\section{Limitations and Ethical Considerations}
In this work, we automatically analyzed social media posts using NLP technology like sentiment and emotion classifiers and machine translation systems, which are error prone. Our selection of tweets was biased via the employed hashtag, i.e., \#ChatGPT. Our human annotation was in some cases subjective and not without disagreements among annotators. Our selection of SemanticScholar papers was determined by the search results of SemanticScholar, which seem non-deterministic. Our search for papers on Arxiv was restricted to mentions in the abstracts and titles of papers and our annotations were only based on titles and abstracts. 
We freely used ChatGPT as an aide throughout the writing process. All errors, however, are our own. 

\section{Acknowledgments}
Ran Zhang, Christoph Leiter, Daniil Larionov are financed by the BMBF project ``Metrics4NLG''. Steffen Eger is financed by DFG Heisenberg grant EG 375/5–1. 
This paper was written while on a retreat in the Austrian mountains in the small village of Hinterriß. 

\bibliographystyle{acl_natbib}
\bibliography{literature}

\clearpage

\end{document}